\pdfoutput=1

\documentclass[11pt]{article}

\usepackage[final]{acl}

\usepackage{times}
\usepackage{latexsym}

\usepackage[T1]{fontenc}

\usepackage[utf8]{inputenc}

\usepackage{microtype}

\usepackage{inconsolata}

\usepackage{graphicx}

\usepackage{comment}
\usepackage{colortbl} 
\usepackage{multirow}

\usepackage{listings}
\usepackage{tcolorbox}

\usepackage{definitions}

\usepackage{todonotes}

\usepackage{amsmath}
\usepackage{xspace}
\usepackage{stfloats}
\usepackage{enumitem}
\usepackage{float}

\usepackage{amssymb}
\usepackage{booktabs}

\title{Diverse LLMs or Diverse Question Interpretations? \\ That is the Ensembling Question}

\author{
  Rafael Rosales 
  \\
  Intel Labs
  \\
  Munich, Germany
  \\
  \texttt{rafael.rosales@intel.com}
  \And
  Santiago Miret
  \\
  Intel Labs
  \\
  Santa Clara, U.S.A.
  \\
  \texttt{santiago.miret@intel.com}
}

\date{}

\begin{document}
\maketitle
\begin{abstract}
Effectively leveraging diversity has been shown to improve performance for various machine learning models, including large language models (LLMs).
However, determining the most effective way of using diversity remains a challenge.
In this work, we compare two diversity approaches for answering binary questions using LLMs: model diversity, which relies on multiple models answering the same question, and question interpretation diversity, which relies on using the same model to answer the same question framed in different ways.
For both cases, we apply majority voting as the ensemble consensus heuristic to determine the final answer.
Our experiments on boolq, strategyqa, and pubmedqa show that question interpretation diversity consistently leads to better ensemble accuracy compared to model diversity.
Furthermore, our analysis of GPT and LLaMa shows that model diversity typically produces results between the best and the worst ensemble members without clear improvement.

\end{abstract}

\section{Introduction and Related Work}
\label{l:intro}

A common goal of embedding diversity into a design is to create a system from multiple components that fail in complementary ways leading to more robustness and better performance~\cite{AvizienisK84}. 
The concept of design diversity has also been applied to machine learning in various ways~\cite{GongZH19, tang2006analysis, pugh2016quality, RosalesMP23, khadka2021optimizing, miret2022neuroevolution}, including model ensembling~\cite{Breiman96b, wang-etal-2017-sogou, chan21, pitis2023boostprompt, jiang-etal-2023-llm} where inference is performed on multiple models and the outcome is a function of the individual outputs, such as majority voting.
For language modeling, Mixture-of-Experts~\cite{12moe, jiang2024mixtralexperts, yuksel2012twenty, ravaut-etal-2022-summareranker} is another application of design diversity, where the architecture of a single LLM consists of separate expert subnets.
Diversity has also been investigated within In-Context Learning (ICL) techniques~\cite{dong2024surveyicl}, such as in the use of diverse reasoning paths~\cite{Li:23, wang2022self}, Tree of Thoughts~\cite{yao2024tree}, results fusion~\cite{WenhaoXJ23}, and also for domain-specific language modeling applications \cite{song-etal-2023-honeybee, lala2023paperqa, gonzales2024evaluating}.
Useful diversity, however, is not straightforward to obtain, as greater disagreement does not necessarily lead to better performance.
A system with useful diversity consists of components that have failure diversity, i.e., when failures are not positively correlated, such that when one component fails the others can compensate for it.
\citet{LM:89} showed that systems obtained from methodologies that follow \textit{diverse design decisions} can yield negatively correlated failures, whereas merely independently developed systems will (on average) not achieve negatively correlated failures.

Ensembles of independently developed LLMs may suffer from shared failure patterns, limiting gains in question-answering accuracy due to positively correlated errors.
Several methods address LLM selection for diverse ensembles.
\citet{jiang-etal-2023-llm} propose pairwise comparisons based on a scoring metric, followed by generative fusion of top-ranked models.
This incurs high cost due to scoring every model pair at inference.
\citet{lv-etal-2024-urg} avoid this by combining ranking and fusion in a single architecture, but this increases complexity and requires training a model to jointly learn both tasks.
Instead of relying on costly training or fine-tuning to enforce methodological diversity across LLMs, we propose leveraging inference-time diversity inherent to question answering—specifically, question interpretation.
Even binary questions in popular datasets such as BoolQ~\cite{clark2019boolq} often exhibit ambiguity; in a random sample, we identified 33\% as ambiguous (see Appendix~\ref{sec:appendix-manual}).
By deliberately varying how a question is interpreted, we enforce diversity as a design choice.
This strategy also requires only a single LLM, streamlining development by enabling focused tuning and prompt engineering.

In this work, we ask the following \researchquestionNOINDEX{rq-which-diversity}{"Can diversity in question interpretation serve as an effective alternative to model diversity?"}.
As shown in Section~\ref{l:results}, interpretation diversity consistently improves performance on binary QA tasks across ChatGPT- and LLaMA-based ensembles, often surpassing model diversity.

\begin{figure}[t]
  \includegraphics[width=\columnwidth]{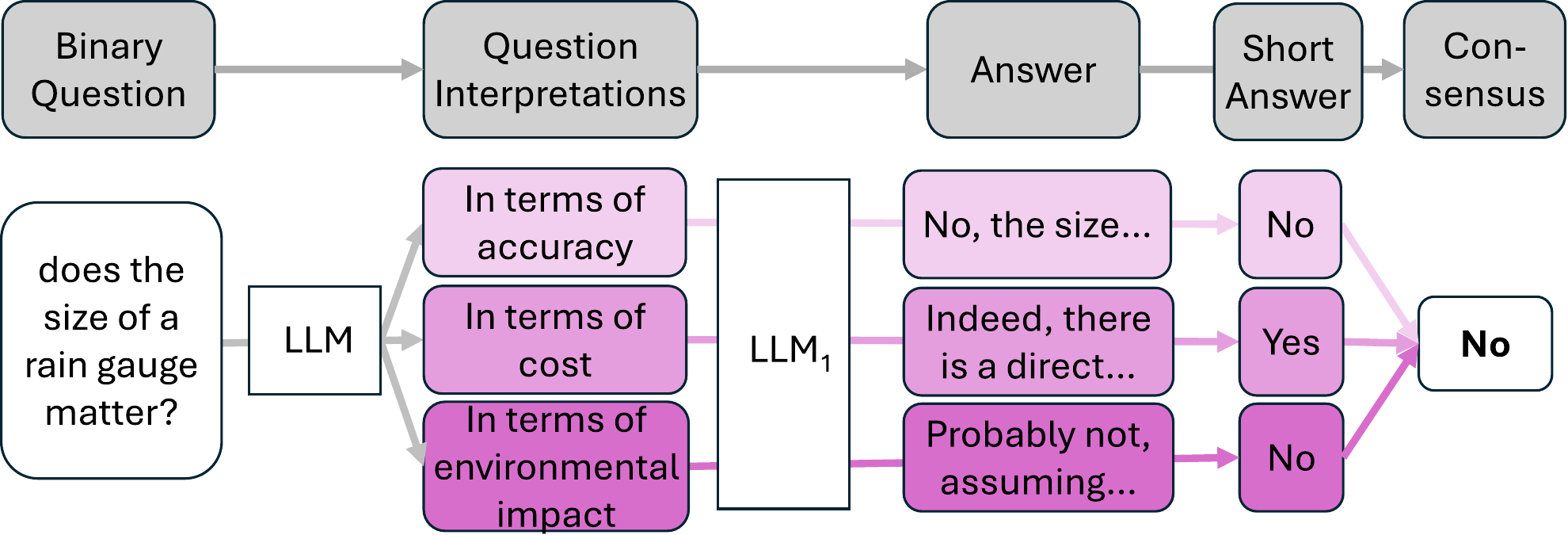}
  \caption{Interpretation diversity: An LLM generates three interpretations of a question, answers each, and the final answer is chosen by majority voting.}
  \label{fig:sol_interpretations}
  \vspace{-1.2em}
\end{figure}

\begin{figure}[t]
  \includegraphics[width=\columnwidth]{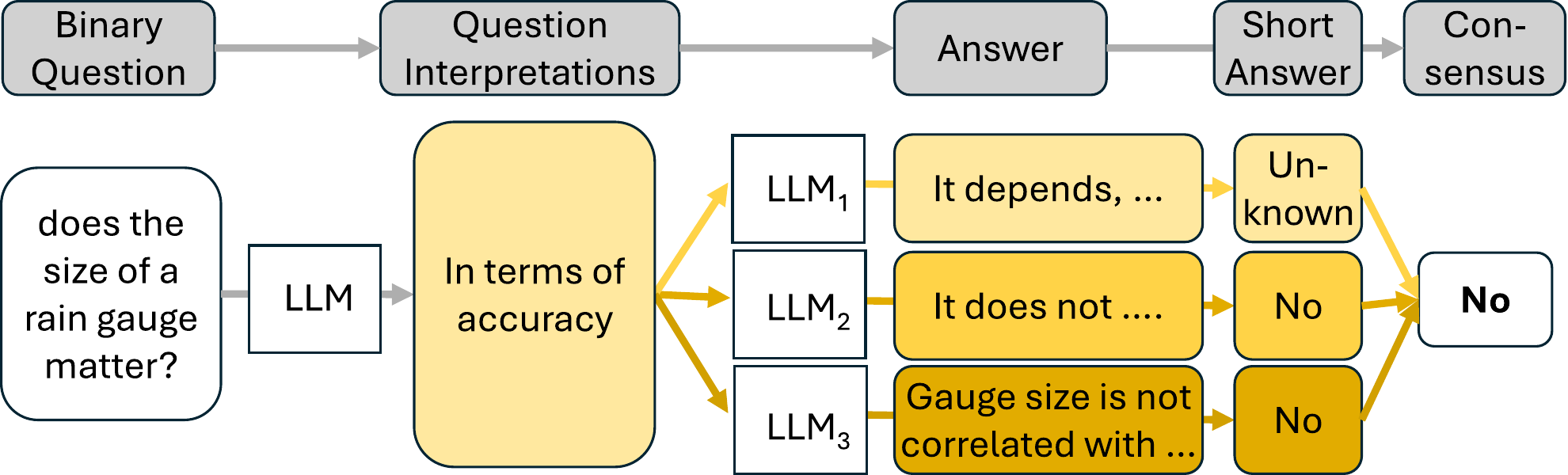}
  \caption{Model diversity: Three different LLMs answer a single interpretation of a question. Majority voting is then performed on the answer's summary.}
  \label{fig:sol_models}
  \vspace{-1.2em}
\end{figure}

Figures~\ref{fig:sol_interpretations} and~\ref{fig:sol_models} illustrate the two ensemble strategies.
In the interpretation diversity approach (Fig.\ref{fig:sol_interpretations}), a fixed LLM generates multiple interpretations of a question, then answers each one. A majority vote is cast over the resulting answers from the same model, ensuring that only interpretation—and not model—diversity is at play.
In contrast, model diversity (Fig.\ref{fig:sol_models}) holds the interpretation constant and aggregates answers from multiple distinct LLMs via majority voting.
Before voting, we apply an answer validation step involving keyword matching and pattern recognition supported in prior work alongside sentiment and likelihood analysis~\cite{24Dimitriadis, 07ieeeAnswerExtract, cole2023selectively}.

To the best of our knowledge, our paper is the first work to directly compare model diversity and question interpretation diversity in LLM-based ensembles.
Nonetheless, related works have studied approaches to reduce question ambiguity, which is related to question interpretation diversity, such as through the use of database of unambiguous questions~\cite{sun23ambigdatabase}, with retrieval augmentation~\cite{kim23ambig, behnamghader2023can}, by asking follow-up questions~\cite{tix24ask}, and through sampling repetition~\cite{cole2023selectively}.
Works such as~\cite{rosenthal21qexpl} explore the need to provide more context to the answers to achieve higher quality answers and enable explainability as a proxy for trustworthiness.

The rest of the paper is structured as follows:
Section~\ref{l:methodology} describes the methodology used,
Section~\ref{l:experiments} outlines the experimental setup,
Section~\ref{l:results} presents the obtained results, and
Section~\ref{l:conclusions} summarizes our findings and provides concluding remarks.

\section{Methodology}
\label{l:methodology}

\paragraph{Metrics}
\label{l:metrics}

We define heuristics and metrics to compare both diversity methods for binary questions.

An LLM model $\llm{i}$ fed with a prompt $\prompt{j}$ produces an answer denoted $\rawAnswer{i}{j} = \llm{i}(\prompt{j})$.
This answer is sent to an answer extraction heuristic $\heuristic{k}$ that produces a summarized answer $\summaryAnswer{i}{j}{k} = \heuristic{k}(\rawAnswer{i}{j}), \; \heuristic{k}: \rawAnswer{i}{j} \rightarrow \summaryAnswers = \{ \texttt{yes}, \texttt{no}, \texttt{none}, \texttt{depends}, \texttt{unclear} \}$.
The set of possible summarized answers is larger than a simple binary `yes' or `no', as LLMs sometimes do not limit their answers to such absolute categories.

The binary \emph{correctness score} is then computed on the extracted summarized answer $\correctness{i}{j}{k} = C(\summaryAnswer{i}{j}{k}), \; C: \summaryAnswers \rightarrow \{ \texttt{correct}, \texttt{incorrect} \}$.

\paragraph{Answer extraction heuristics} We now present two answer extraction heuristics $E^k$ used for answer validation, i.e., how do we extract a `yes' or `no' answer from any possible LLM response $\rawAnswer{i}{j}$.
Next, we show how we qualify the extracted answer as correct or incorrect.

\subparagraph{Match First} \matchFirst\unskip$(\rawAnswer{i}{j})$. 
This heuristic looks for the first case-insensitive pattern match from a set of typical positive: \matchRSet{yes, possibly yes, possible, likely yes}, and negative answers: \matchRSet{no, possibly no, possible, unlikely, likely no}.
If no match is found, the extracted answer is \texttt{`none'}.

\subparagraph{Match Pattern} \matchPattern\unskip$(\rawAnswer{i}{j})$.
This heuristic consecutively attempts to match a reply from the following manually selected regular expressions: \matchRPattern{<answer>(.*?)<\textbackslash/answer>} $\succ$ \matchRPattern{Answer:(.*?).} $\succ$ \matchRPattern{The answer is (.*?).} $\succ$ \matchFirst.
If no pattern matches, \texttt{`none'} is returned.

\paragraph{Answer validation}
\label{sec:val-metrics}
Once the summary response $\summaryAnswer{i}{j}{k}$ of prompt $\prompt{j}$ is obtained, the correctness score $C$ is compared against the ground truth answer $\gt{j}$ computed as follows:

\[
C(\summaryAnswer{i}{j}{k})
\begin{cases}
correct & \text{if} \; \summaryAnswer{i}{j}{k} = \gt{j} \\
incorrect & \text{if} \; \summaryAnswer{i}{j}{k} \neq \gt{j}
\end{cases}
\] 

Note that the range of $\summaryAnswer{i}{j}{k}$ can contain more elements than the range of the ground truth $\gt{j}$, as many binary question datasets provide only a `yes' or a `no'.
This means, that this metric would always penalize ill-posed questions that may be ambiguous and do not have a clear binary answer.

To evaluate an entire dataset, we employ \emph{Accuracy}: the arithmetic mean of the correctness metric.

\section{Experiments}
\label{l:experiments}

\paragraph{Models}
Two categories of LLMs are evaluated: commercial models based on OpenAI GPT series~\cite{achiam2023gpt}: gpt35turbo, gpt35-i, gpt4; and open weight models based on LLaMA~\cite{touvron2023llama, dubey2024llama}: llama2, llama3, and llama3-i.
The ``i'' variants are fine-tuned "instruct" versions of the LLM tailored to provide an answer to the prompt instruction.
Detailed version information is shown in Appendix~\ref{sec:appendix-models}.

\paragraph{Datasets}
Three datasets of binary questions are extracted from the following benchmarks: a) boolq~\cite{clark2019boolq} b) strategyqa~\cite{geva2021did}, and c) pubmedqa~\cite{jin2019pubmedqa}.
Further details are outlined in Appendix \ref{sec:appendix-datasets}.

\paragraph{Prompting} 

To produce differing interpretations of questions without introducing model variability (see Figure~\ref{fig:sol_interpretations}), we used a single model (gpt35) alongside the prompt template $\prompt{multi-step1}$, as detailed in Appendix~\ref{sec:appendix-prompts}, and each independent model subsequently provides answers for the set of re-interpreted questions.
This is done via the $\prompt{multi-step2}$ prompt, shown in Appendix~\ref{sec:appendix-prompts}, which instructs to add relevant facts to the semantic interpretation of the question before answering.

\paragraph{Majority Voting}

From the responses of the six models to all question interpretations, two different classes of ensembles are created: 
\begin{itemize}
    \item Across interpretations---Ensemble of answers to three different interpretations from a single model. Example: \{gpt35t::interpret1, gpt35t::interpret2, gpt35t::interpret3\}
    \item Across models---Ensemble of three different models responding to the same interpretation. Example: \{gpt35t::interpret1, gpt35t-i::interpret1, gpt4::interpret1\}.
\end{itemize}
Each ensemble provides a single answer through a majority voting of their members:
\[
\majVoting = \arg\max_{a \in \summaryAnswers} \sum_{n=1}^{N} \mathbb{I}(\heuristic{k}_n(\rawAnswer{i}{j}) = a)
\]
where:
\(\majVoting\) is the majority voted answer,
\(\summaryAnswers\) is the set of valid summarized answers,
\(N\) is the number of ensemble members,
$\heuristic{k}_n(\rawAnswer{i}{j})$ is the summarized answer of the \(n\)-th member according to heuristic $\heuristic{k}$ for model answer $\rawAnswer{i}{j}$,
$\mathbb{I}$ is an indicator function that equals 1 if $\heuristic{k}_n(x) = a$ and 0 otherwise.

\section{Results}
\label{l:results}

Dataset-wise accuracy for answers to a) direct questions (baseline), b) interpreted questions, and c) voting variants is shown in Table~\ref{t:results}.

\begin{table*}[h!]
\footnotesize
\centering
\begin{tabular}{cl|r|r|r|r|c|c}
\hline
\textbf{Task} & \textbf{Model} & \textbf{Baseline} & \textbf{Interp. 1} & \textbf{Interp. 2} & \textbf{Interp. 3} & \textbf{Vote-Interp.} & \textbf{Vote-Model} \\
\hline
\multirow{6}{*}{boolq}
  & gpt3.5   & 54.89 & 75.23 & 65.26 & 60.09 & 72.63 &  \\
  & gpt3.5-i & 60.58 & 66.79 & 65.69 & 64.50 & 66.27 & 77.31      \\
  & gpt4     & \textbf{80.43} & 79.63 & 71.31 & 69.79 & \underline{77.89} &       \\ \cline{2-8}
  & llama2   & 46.54 & 9.42  & 8.04  & 9.66  & 39.39 &  \\
  & llama3i  & 57.06 & 47.98 & 40.58 & 37.58 & \underline{\textbf{61.35}} & 51.16      \\
  & llama3   & 51.19 & 35.41 & 32.91 & 31.90 & 49.48 &       \\
\hline
\multirow{6}{*}{pubmedqa}
  & gpt3.5   & 29.21 & 86.52 & 67.87 & 48.31 & 78.43 &  \\
  & gpt3.5-i & 42.02 & 99.10 & 96.18 & 98.65 & \underline{\textbf{99.78}} & 92.81       \\
  & gpt4     & 68.54 & 77.75 & 77.30 & 79.10 & 81.80 &       \\ \cline{2-8}
  & llama2   & 41.80 & 15.06 & 7.87  & 14.16 & 4.72  &  \\
  & llama3i  & 46.97 & 57.08 & 48.31 & 44.49 & \underline{\textbf{73.03}} & 41.80      \\
  & llama3   & 55.06 & 43.37 & 38.65 & 41.12 & 38.65 &       \\
\hline
\multirow{6}{*}{strategyqa}
  & gpt3.5   & 58.86 & 65.68 & 58.12 & 54.63 & 64.89 &  \\
  & gpt3.5-i & 60.17 & 52.58 & 52.18 & 53.10 & 52.27 & 68.03      \\
  & gpt4     & \textbf{74.89} & 70.61 & 62.58 & 65.72 & \underline{69.69} &       \\ \cline{2-8}
  & llama2   & 50.74 & 9.13  & 6.55  & 8.08  & 53.76 &  \\
  & llama3i  & \textbf{56.77} & 40.96 & 39.52 & 34.76 & \underline{55.46} & 54.15      \\
  & llama3   & 52.62 & 31.88 & 29.65 & 27.21 & 51.62 &       \\
\hline
\end{tabular}
\caption{Average scores per dataset for models answering the baseline, individual interpretations, and voting variants. Bold indicates the best in each dataset-model group; underline marks the better of interpretation vs. model voting.}
\label{t:results}
\end{table*}

\subsection{Voting across diverse interpretations}
\label{l:voting}

\begin{figure}[htbp]
  \includegraphics[width=0.49\columnwidth]{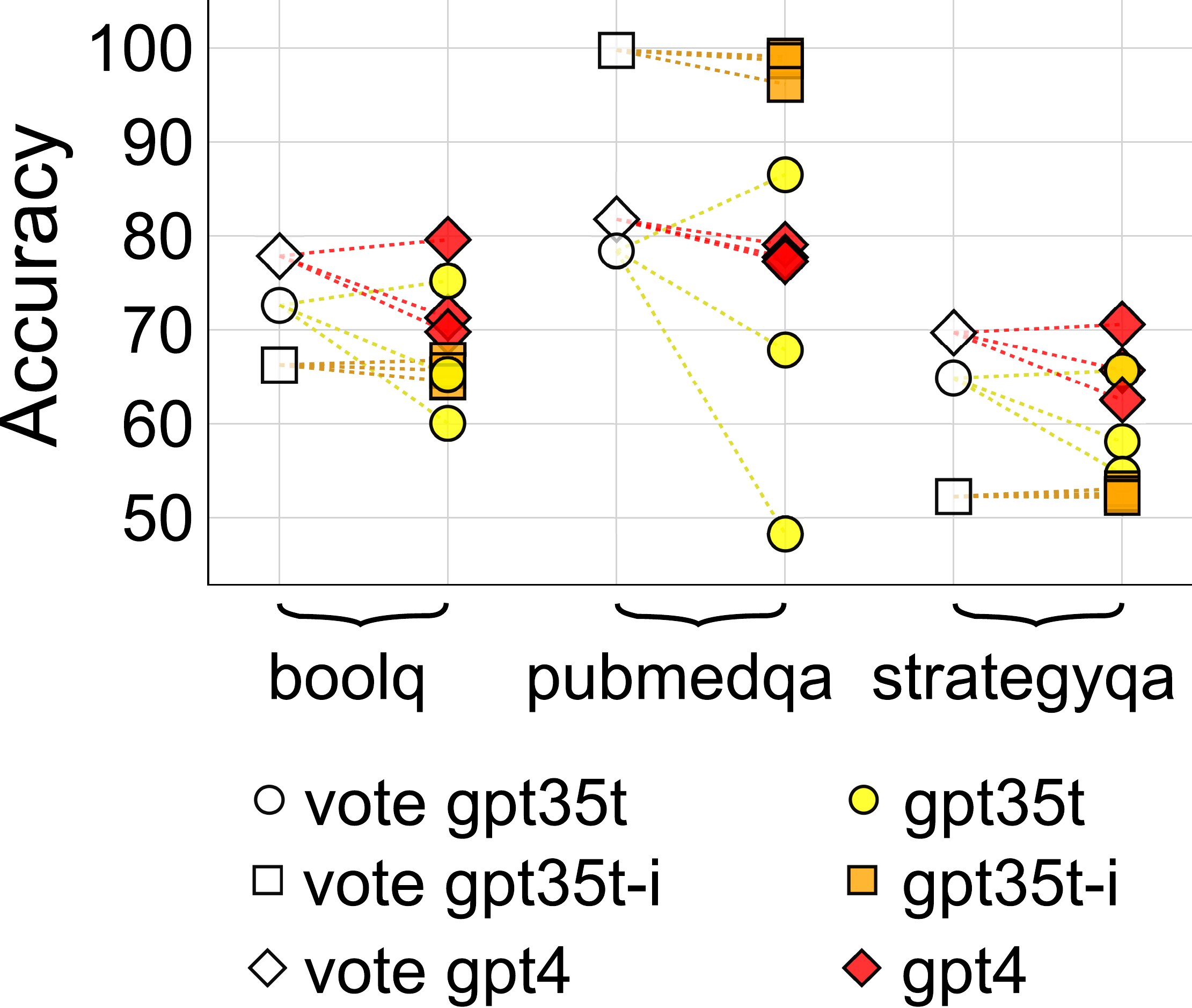}
  \hfill
  \includegraphics[width=0.49\columnwidth]{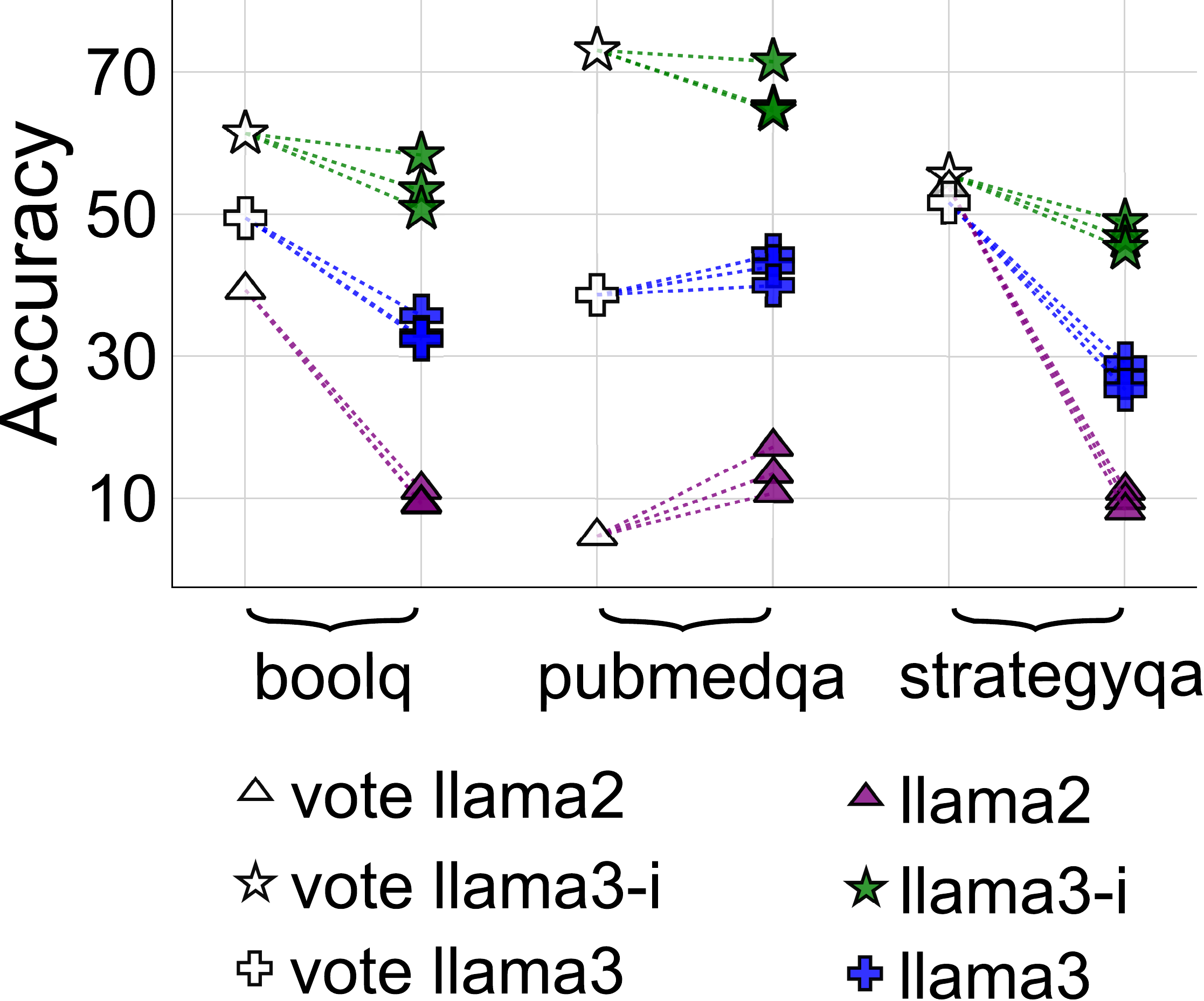}
  \caption{Voting across interpretations: Majority voting accuracy (white) vs. individual interpretation accuracy (color). Left: ChatGPT models; Right: LLaMA models.}
  \label{fig:voting_interpret}
  \vspace{-0.7em}
\end{figure}

Figure~\ref{fig:voting_interpret} shows accuracy scores for interpretation diversity ensembles, with dotted lines linking each interpretation’s score to the ensemble score.

Figure~\ref{fig:voting_interpret} shows that \textit{when voting across interpretations, the ensemble score is above or close to the best member score most of the time}, except the untuned llama2 and llama3 models in the pubmedqa dataset.
Furthermore, the \textit{improvement of the ensemble is significant for the LLaMA models in boolq and strategyqa datasets}, and a slight improvement on the llama3-i model on pubmedqa.

\subsection{Voting across diverse models}
\label{l:voting_models}

\begin{figure}[h]
  \includegraphics[width=0.49\columnwidth]{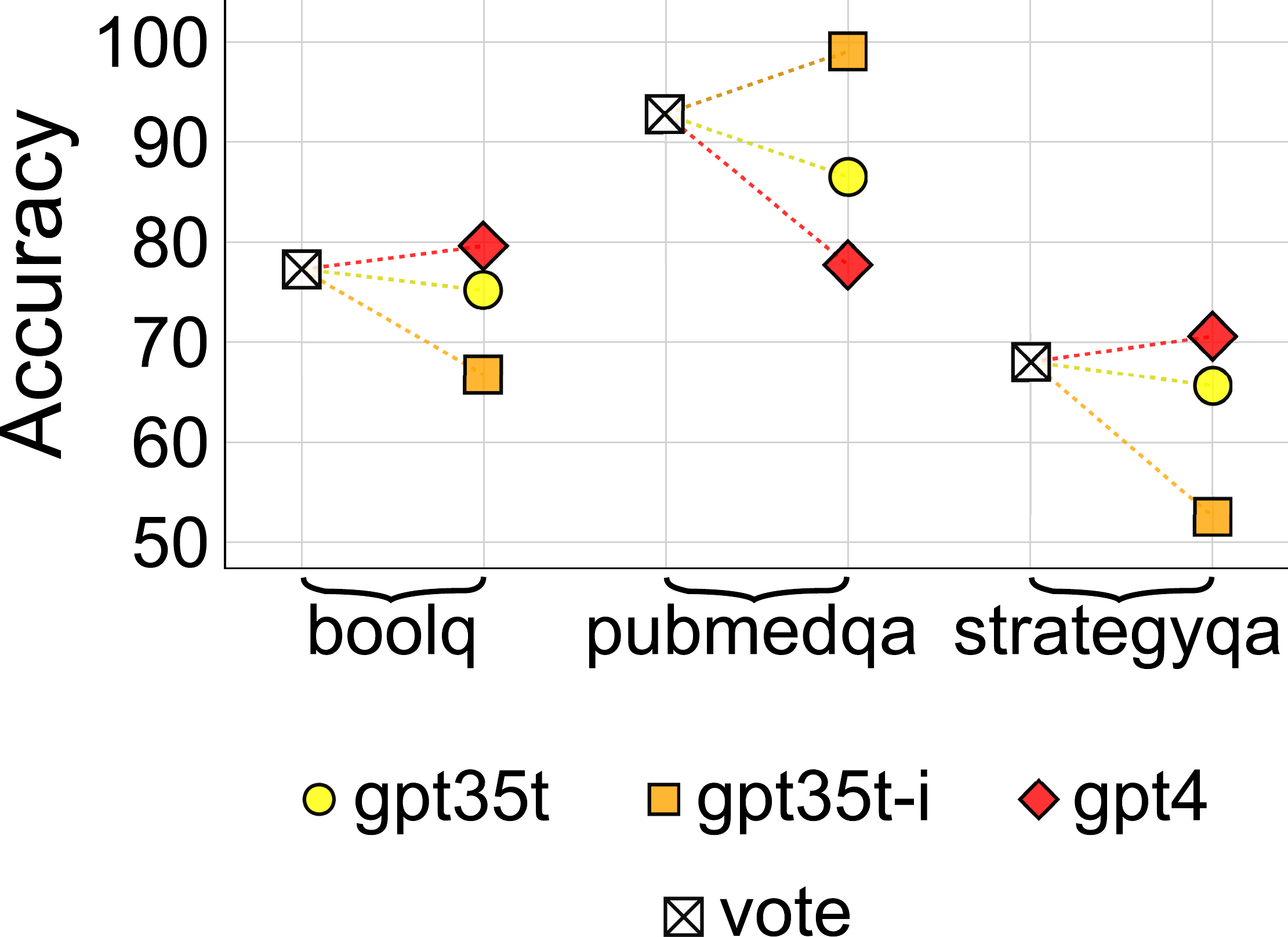}
  \hfill
  \includegraphics[width=0.49\columnwidth]{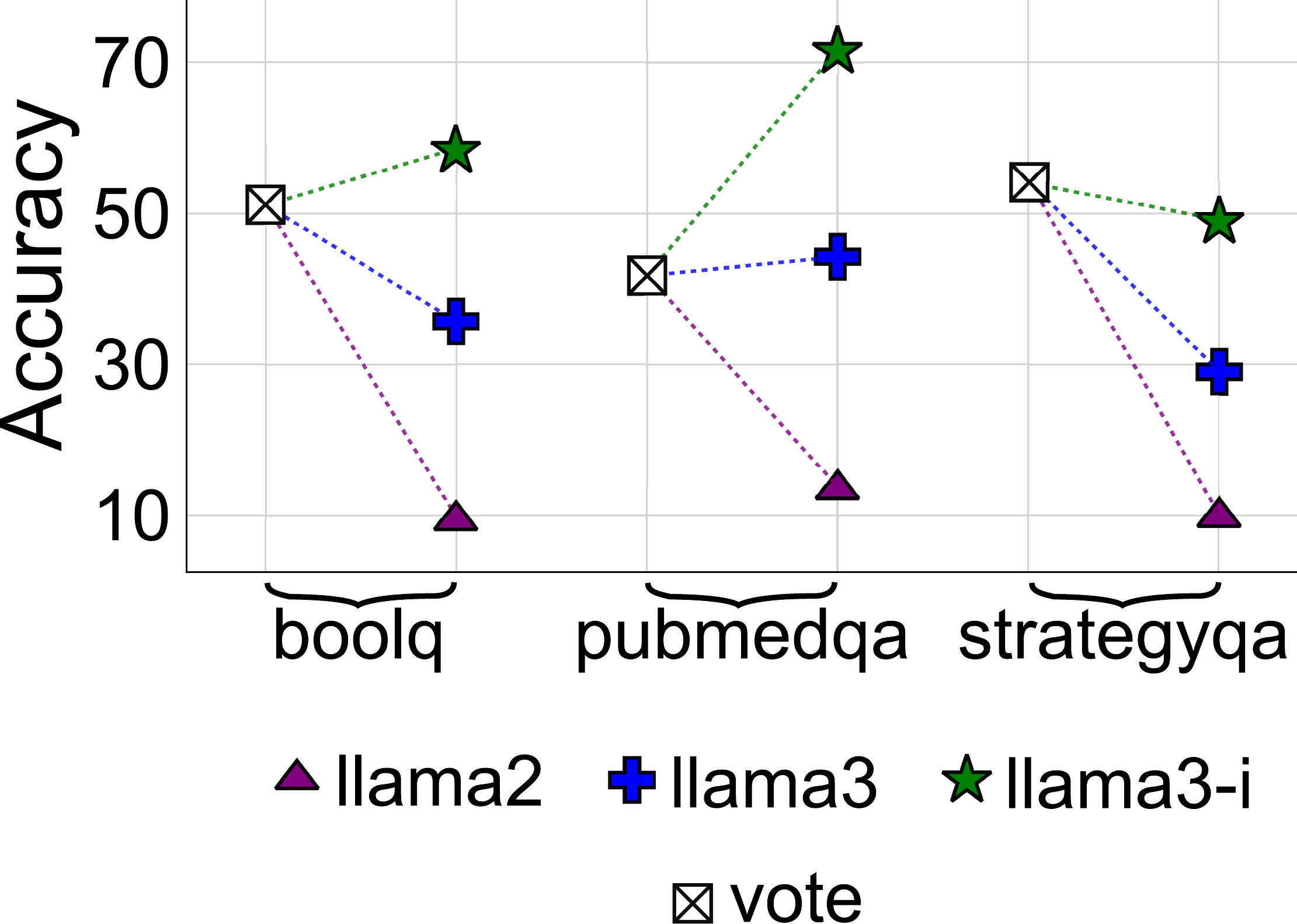}
  \caption{Voting across models: Majority voting accuracy (white crossed rectangle). Left: ChatGPT ensemble; Right: LLaMA ensemble.}
  \label{fig:voting_models_multistep}
  \vspace{-1.2em}
\end{figure}

Figure~\ref{fig:voting_models_multistep} shows GPT (left) and LLaMA (right) ensemble accuracy compared to their individual models across the three datasets.

The results of Figure~\ref{fig:voting_models_multistep} show that \textit{with model diversity, the ensemble scores mostly fall within the range defined by the maximum and minimum scores of the ensemble members}.

\subsection{Interpretation diversity vs model diversity}

\begin{figure}[h]
  \includegraphics[width=\columnwidth]{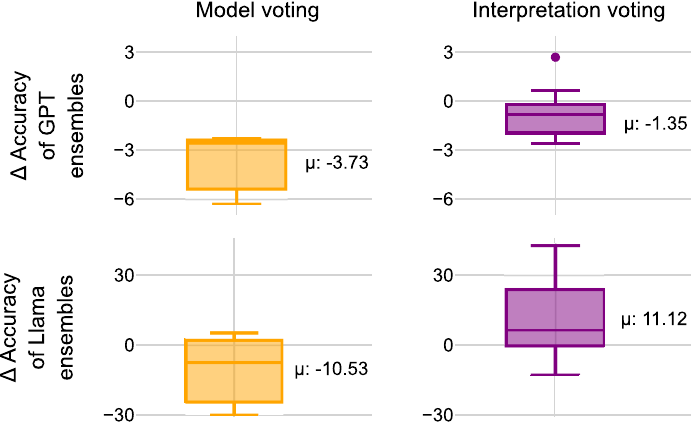}
  \caption{Accuracy delta between the ensemble vote and its best individual model. Interpretation voting (right) beats model voting (left) for both ChatGPT and LLaMA.}
  \label{fig:deltas}
  \vspace{-1.2em}
\end{figure}

To compare the results of the two diversity approaches, the distribution of the accuracy differences of the ensembles relative to the best individual member, is shown in Figure~\ref{fig:deltas} for the ensembles based on ChatGPT models and for those based on LLaMA models.

In Figure~\ref{fig:deltas}, we observe that for both the ChatGPT and LLaMA ensembles, the \textit{interpretation diversity approach is superior compared to ensemble diversity}, as it consistently results in a better accuracy difference.
The mean accuracy differences to the best individual member of model diversity are -3.73 and -10.53 for ChatGPT and LLaMA ensembles respectively, while for interpretation diversity the means are -1.35 and 11.12, revealing that interpretation diversity results are closer to the best individual ensemble member.
In addition, it can be observed that \textit{interpretation diversity provides a significant improvement with respect to the best ensemble member for the case of LLaMA ensembles}.

\section{Conclusions}
\label{l:conclusions}

In closing, our results show that, for the tested models and datasets, question interpretation diversity leads to better ensemble accuracy than model diversity.
The accuracy improvement distributions of question interpretation voting consistently exceed those of model diversity, with minimal overlap.
This advantage is particularly pronounced in the case of the LLaMA-based ensembles. 
In contrast, model diversity typically produces results that fall between the best and worst ensemble members.

Future work could explore enhancements through more sophisticated heuristics and ensemble algorithms, such as weighting interpretation validity to improve consensus or how diversity can be applied to reasoning models \cite{luo2024improve}.

\section{Limitations and Ethical Concerns}

\subsection{Limitations}

\paragraph{Datasets}The work at hand has been evaluated on the task of answering binary questions from selected established benchmarks.
These datasets do not capture all types of question-answering tasks LLMs may encounter and contain some additional limitations.
Although the number of `yes' and `no' ground truth answers is balanced, and the domains covered are broad, manual inspection reveals that a human would not necessarily agree with the ground truth.
Even though our results are limited to the evaluated datasets, they provide a reference point for comparing diversity methods.

\paragraph{Binary vs open-ended questions}
In order to evaluate the diversity of question interpretation to open-ended questions it would be necessary to extend the answer validation step.
If the ground truth answers are not limited to `yes' or `no', it becomes significantly harder to identify if a model has correctly answered as the number of potentially correct answers is not bounded.
Nevertheless, validation heuristics, such as semantic similarity based on text embeddings could be applied, to estimate if an answer is closer to the ground truth.

\paragraph{LLMs tested}Our experiments focused on only two categories of LLMs: ChatGPT- and LLaMA-based. 
We chose to evaluate a single class of commercial LLMs and a single class of open-weight LLMs as representatives of black-box and open foundational models.
The evaluation of further models would further solidify the strength of the results.
Furthermore, more sophisticated ensembles could be evaluated, such as comparing fine-tuned models vs foundational models, as well as heterogeneous ensembles.

\paragraph{Performance}The use of ensembling introduces an overhead that linearly scales with the number of ensemble members used.
A majority vote of three members would require $3\times$ the number of inferences.
Furthermore, the creation of an extra step to provide an explicit semantic interpretation of the question introduces a latency overhead, as the ensemble members would need to wait for the interpretation to be generated.

\paragraph{Combination with other techniques}The investigated diversity approaches could be combined with existing and orthogonal state-of-the-art techniques for LLMs such as fine-tuning from human feedback~\cite{NEURIPS2022_rlhf}, fine-tuning for a domain application~\cite{han2024peftllm}, ICL prompting engineering techniques~\cite{dong2024surveyicl}, as well as Retrieval Augmented Generation (RAG)~\cite{NEURIPS2020_rag, behnamghader2023can}.
Such techniques can be applied to individual LLM inference calls.
We leave the evaluation of these techniques, and their potential combinations, to future work.

\paragraph{Scalability}
It would be valuable to examine whether increasing the size of a model ensemble eventually yields greater benefits than adding additional interpretations. This remains unclear, as the number of valid interpretations may be finite, yet research also suggests that the advantages of ensembling multiple models eventually plateau.

\subsection{Ethical Concerns}

In general, diversity can be a positive contributor to reducing bias.
When applied to identify different question interpretations, this technique could be helpful to address inherent biases in the question itself by providing a set of different semantic interpretation possibilities.
However, the consensus heuristic would potentially introduce a point of concern.
Depending on what heuristic is chosen, the result could again be biased.
Majority voting itself is an imperfect voting mechanism, and majority voting without weights gives every opinion the same weight, independent of any factor.

\bibliography{references}

\appendix

\section{Model details}
\label{sec:appendix-models}

The number of parameters of the LLaMA models can be identified in the model ID of Table~\ref{tab:llms}.

\begin{table}[h]
\centering
\setlength{\tabcolsep}{2pt} 
\small
\begin{tabular}{|c|c|c|}
\hline
id & LLM & Source\\
\hline
gpt35t & gpt3.5-turbo-0301 & \cite{openaiGPT} \\
\hline
gp35t-i & gpt3.5-turbo instruct-0914 & \cite{openaiGPT} \\
\hline
gpt4 & gpt4-0125-preview & \cite{openaiGPT} \\
\hline
llama2 & LLaMA-2-7b-chat-hf & \cite{metallama} \\
\hline
llama3 & Meta-LLaMA-3-8B & \cite{metallama} \\
\hline
llama3-i & Meta-LLaMA-3-8B-Instruct & \cite{metallama}\\
\hline
\end{tabular}
\caption{Exact version of tested LLMs}
\label{tab:llms}
\end{table}

GPT-3.5t generated all interpretations to avoid a complex combinatorial space and is closer in scale and behavior to open-source models than GPT-4.

\section{Dataset details}
\label{sec:appendix-datasets}
The datasets used in this study are:
\begin{itemize}
    \item boolq: All 3270 questions of the BoolQ validation dataset~\cite{clark2019boolq}.
    \item strategyqa: All 2290 questions from the StrategyQA train dataset~\cite{geva2021did} and 
    \item pubmedqa: A total of 445 questions from a \%50 split of the PQA-L yes/no questions in the PubMedQA dataset~\cite{jin2019pubmedqa}.
\end{itemize}
BoolQ and StrategyQA are open domain questions and PubMedQA is specific to biomedicine.

\section{Prompt templates}
\label{sec:appendix-prompts}

The prompt templates were manually selected by empirically trying known prompt tactics, such as CoT, adopt a persona, use of delimiters, and multiple steps, on three sample questions that would produce binary responses in the expected format.

Prompt used to genereate question interpretations: $\prompt{multi-step1}$
\begin{promptbox}{Interpretation prompt : $\prompt{multi-step1}$}
 "System: Consider the question: \verb|'|\textcolor{blue}{\{question\}}\verb|'|. How might this question be interpreted in different contexts or with different underlying meanings? Provide at least three distinct semantic interpretations."
\end{promptbox}

Prompt used to answer a question using a specific interpretations: $\prompt{multi-step2}$
\begin{promptbox}{Answer interpretation prompt : $\prompt{multi\text{-}step2}$}
	"System: The original question \verb|'|\textcolor{blue}{\{question\}}\verb|'| has multiple interpretations (indexed from 1 to N). First, I want you to state if Interpretation \verb|'|\textcolor{blue}{\{index\}}.\verb|'| is mentioned in the list or not: \{ \textcolor{blue}{\{interpretations\}} \}. Second, if it is not mentioned reply with \verb|'|<answer>EMPTY</answer>\verb|'| and stop, but otherwise I want you to exactly repeat interpretation \verb|'|\textcolor{blue}{\{index\}}.\verb|'| and then state the most relevant facts to answer this interpretation. Finally, summarize the answer with \verb|'|<answer>Yes</answer>\verb|'| or \verb|'|<answer>No</answer>\verb|'|."
\end{promptbox}

\section{Manual evaluation}
\label{sec:appendix-manual}

To better understand the experiment results, we show some manual validation analysis results.

We sampled ~30 questions with a uniformly distributed probability across the open domain datasets, see Table~\ref{tab:sampleBoolq} and Table~\ref{tab:sampleStrategy} show these questions and the ground truth answers.
Table~\ref{tab:sample_answers_gpt35t} and Table~\ref{tab:sample_answers_llama3i} show the answers of model gpt35t and llama3-i to three different interpretations of one of the boolq questions.

\newcommand{\ambQColor}[0]{gray}

\begin{table}[h]
\centering
\tiny
\begin{tabular}{|p{6.5cm}|p{0.2cm}|}
\hline
 QUESTION & GT \\ \hline
 is angular frequency and angular velocity the same & no \\ \hline
 \cellcolor{\ambQColor}is the singer from blur in the gorillaz & yes \\ \hline
 \cellcolor{\ambQColor}does the size of a rain gauge matter & yes \\ \hline
 is a cougar and a mountain lion the same & yes \\ \hline
 is skyline drive part of the blue ridge parkway & yes \\ \hline
 \cellcolor{\ambQColor}does the jeep commander have a third row & yes \\ \hline
 does average velocity have a direction associated with it & yes \\ \hline
 \cellcolor{\ambQColor}can an x be used as a signature & yes \\ \hline
 \cellcolor{\ambQColor}is there something at the end of ifinity war & yes \\ \hline
 can a terminating decimal be written as a recurring decimal & yes \\ \hline
 did they used to make 1000 dollar bills & yes \\ \hline
 is there a continuation to avengers infinity war & yes \\ \hline
 was the decl. of independence written during the second continental congress & yes \\ \hline
 is there free to air tv in the usa & yes \\ \hline
 are trinidad and tobago in the world cup 2018 & no \\ \hline
 \cellcolor{\ambQColor}is jason in the first friday the 13th & yes \\ \hline
 is a power adapter the same as a charger & no \\ \hline
 \cellcolor{\ambQColor}is there a story to magic the gathering & yes \\ \hline
 does the dad die in train to busan & yes \\ \hline
 is things fall apart based on a true story & no \\ \hline
 did we have red solo cups in 1982 & yes \\ \hline
 \cellcolor{\ambQColor}does the mlb all star game go into extra innings & yes \\ \hline
 was looking for mr goodbar based on a true story & yes \\ \hline
 is eastern kentucky university a division 1 school & yes \\ \hline
 is there a way one can grow more crops from the same land & yes \\ \hline
 is rosso vermouth the same as sweet vermouth & no \\ \hline
 is the international journal of nursing studies peer reviewed & yes \\ \hline
 \cellcolor{\ambQColor}did colombia make it to the round of 16 & yes \\ \hline
 will there be a total bellas season 4 & yes \\ \hline
 \cellcolor{\ambQColor}does the batter have to move out of the way of a pitch & yes \\ \hline
\end{tabular}
\caption{Sampled questions from boolq dataset. Highlighted in gray are questions that can be interpreted in more than one way due to ambiguity or lack of context.}
\label{tab:sampleBoolq}
\end{table}

\begin{table}[h]
\centering
\tiny
\begin{tabular}{|p{6.5cm}|p{0.2cm}|}
\hline
 QUESTION & GT \\ \hline
 \cellcolor{\ambQColor}Is Guitar Hero Beatles inappropriate for a US third grader? & no \\ \hline
 Does a Generation Y member satisfy NYPD police officer age requirement? & yes \\ \hline
 Does Amtrak operate four wheel vehicles? & yes \\ \hline
 \cellcolor{\ambQColor}Is most store bought rice pudding made with brown rice? & no \\ \hline
 Would it be difficult to host Stanley Cup Finals at Rock in Rio? & yes \\ \hline
 Would an owl monkey enjoy a strawberry? & yes \\ \hline
 \cellcolor{\ambQColor}Would the author of Little Women have remembered the ratification of the 13th Amendment?& yes \\ \hline
 Did Mozart ever buy anything from Dolce \& Gabbana? & no \\ \hline
 Are paratroopers good at mountain rescue? & yes \\ \hline
 \cellcolor{\ambQColor}Were French people involved in the American Civil War? & yes \\ \hline
 \cellcolor{\ambQColor}Can you avoid internet trolls on reddit? & no \\ \hline
 \cellcolor{\ambQColor}Do the directors of The Matrix advocate for transgender rights? & yes \\ \hline
 \cellcolor{\ambQColor}Would menu at Chinese Starbucks be familiar to an American? & no \\ \hline
 Do salmon mate in the Caspian Sea? & no \\ \hline
 Can an adult male stand on top Donatello's bronze David and touch the Sistine Chapel ceiling? & no \\ \hline
 Could Al Capone have read works from the Harlem Renaissance? & yes \\ \hline
 Was the subject of Parsifal taken from British folklore? & yes \\ \hline
 Will Dustin Hoffman likely vote for Trump in 2020? & no \\ \hline
 Does the Taco Bell kitchen contain cinnamon? & yes \\ \hline
 Is Issac Newton often associated with a red fruit? & yes \\ \hline
 Are all the elements plants need for photosynthesis present in atmosphere of Mars? & yes \\ \hline
 \cellcolor{\ambQColor}Would Eye surgery on a fly be in vain? & yes \\ \hline
 Can Paprika be made without a dehydrator? & yes \\ \hline
 Would an Alfa Romeo vehicle fit inside a barn? & yes \\ \hline
 Is Alan Alda old enough to have fought in the Vietnam War? & yes \\ \hline
 Was Oscar Wilde's treatment under the law be considered fair in the US now? & no \\ \hline
 Would a Bengal cat be afraid of catching a fish? & no \\ \hline
 Could Edward Snowden have visited the headquarters of United Nations Conference on Trade and Development? & yes \\ \hline
 Will a sloth explode if it's not upside down? & no \\ \hline
\end{tabular}
\caption{Sampled strategyqa dataset questions. Highlighted in gray are questions that can be interpreted in more than one way due to ambiguity or lack of context.}
\label{tab:sampleStrategy}
\end{table}

\begin{figure}[h]
  \includegraphics[width=0.49\columnwidth]{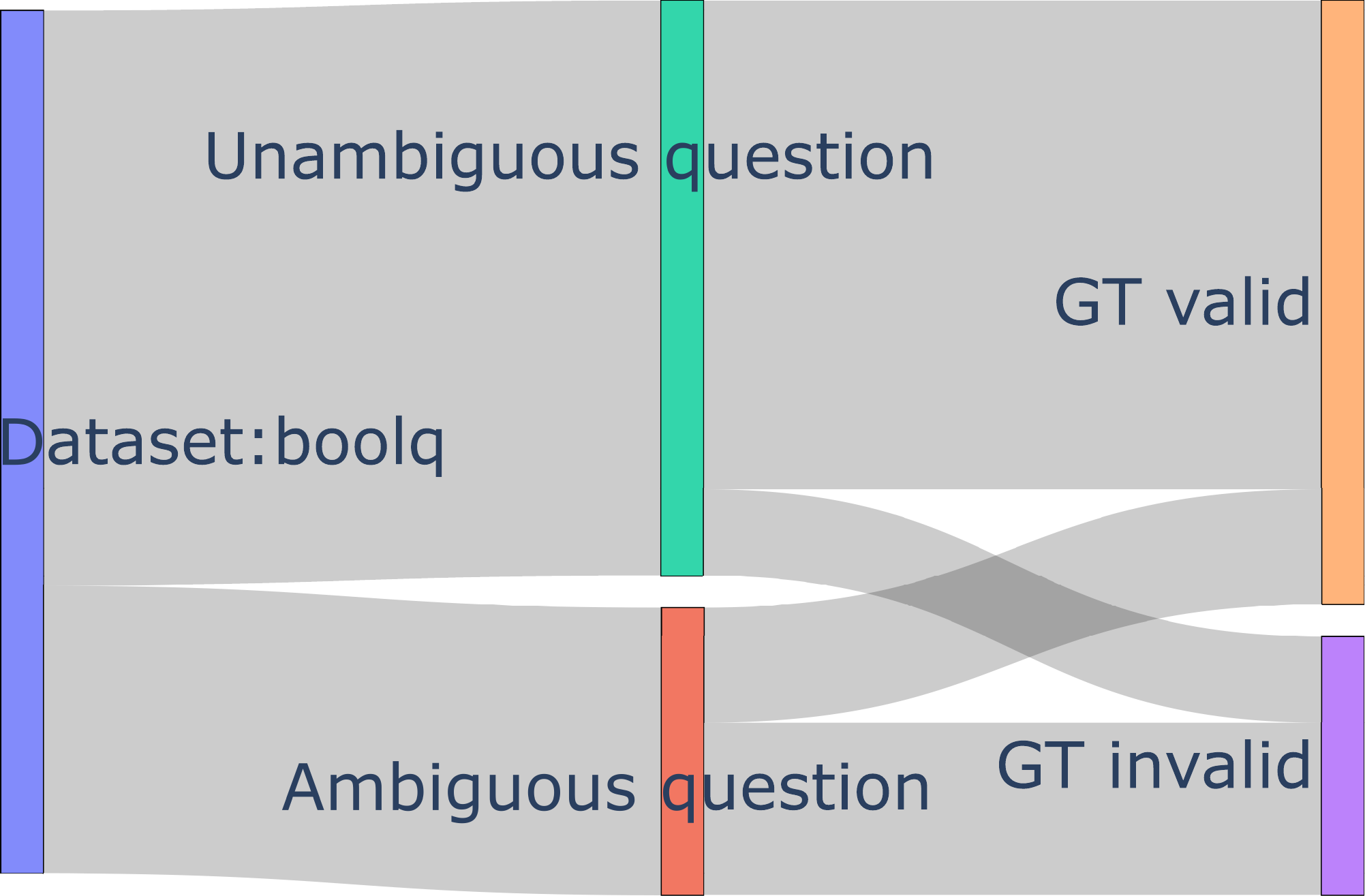}
  \hfill
  \includegraphics[width=0.49\columnwidth]{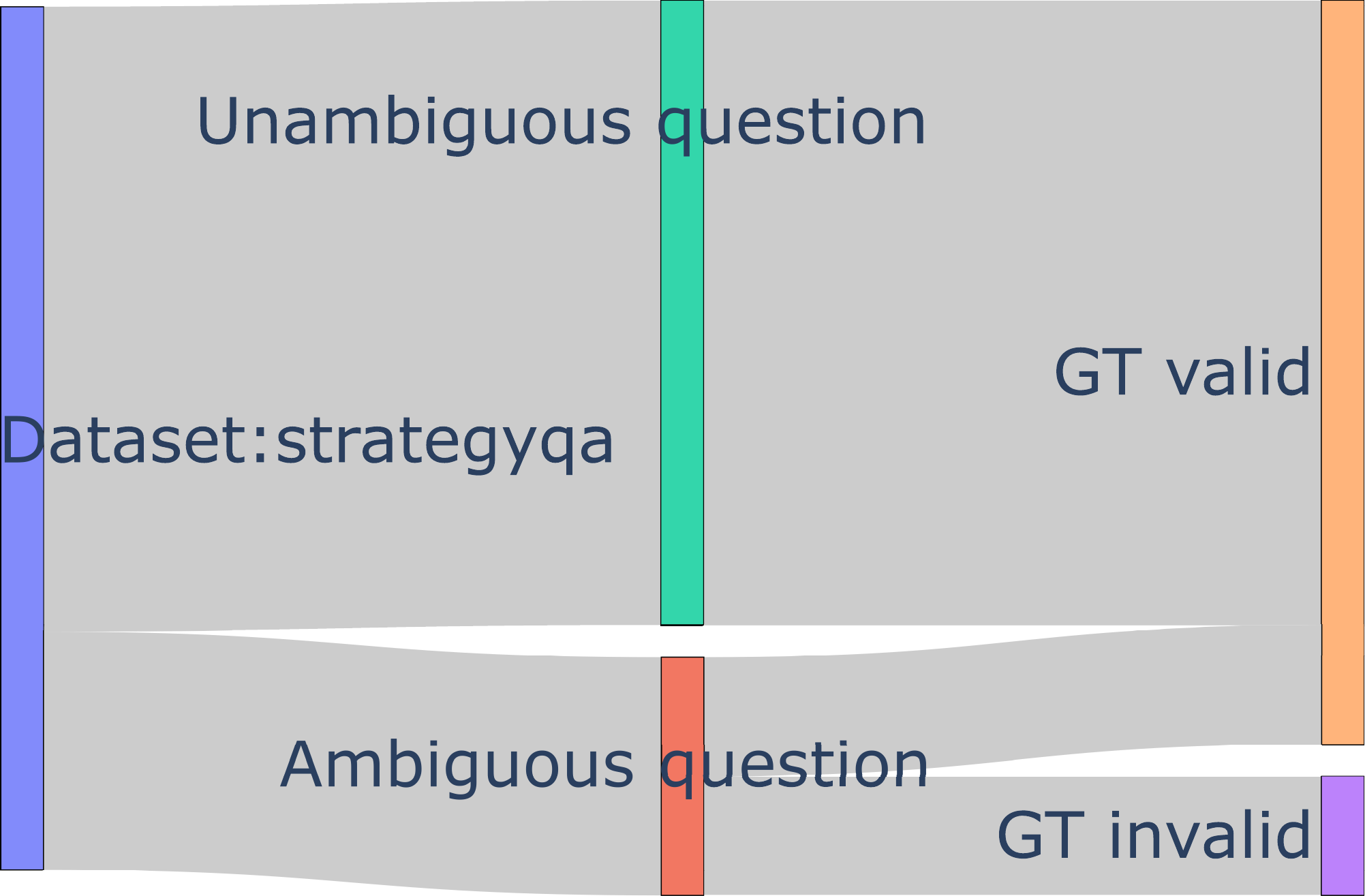}
  \caption{Manual classification of sample questions. Left: boolq. Right: strategyqa.}
  \label{fig:sampling_questions}
\end{figure}

\begin{figure}[h]
  \includegraphics[width=0.49\columnwidth]{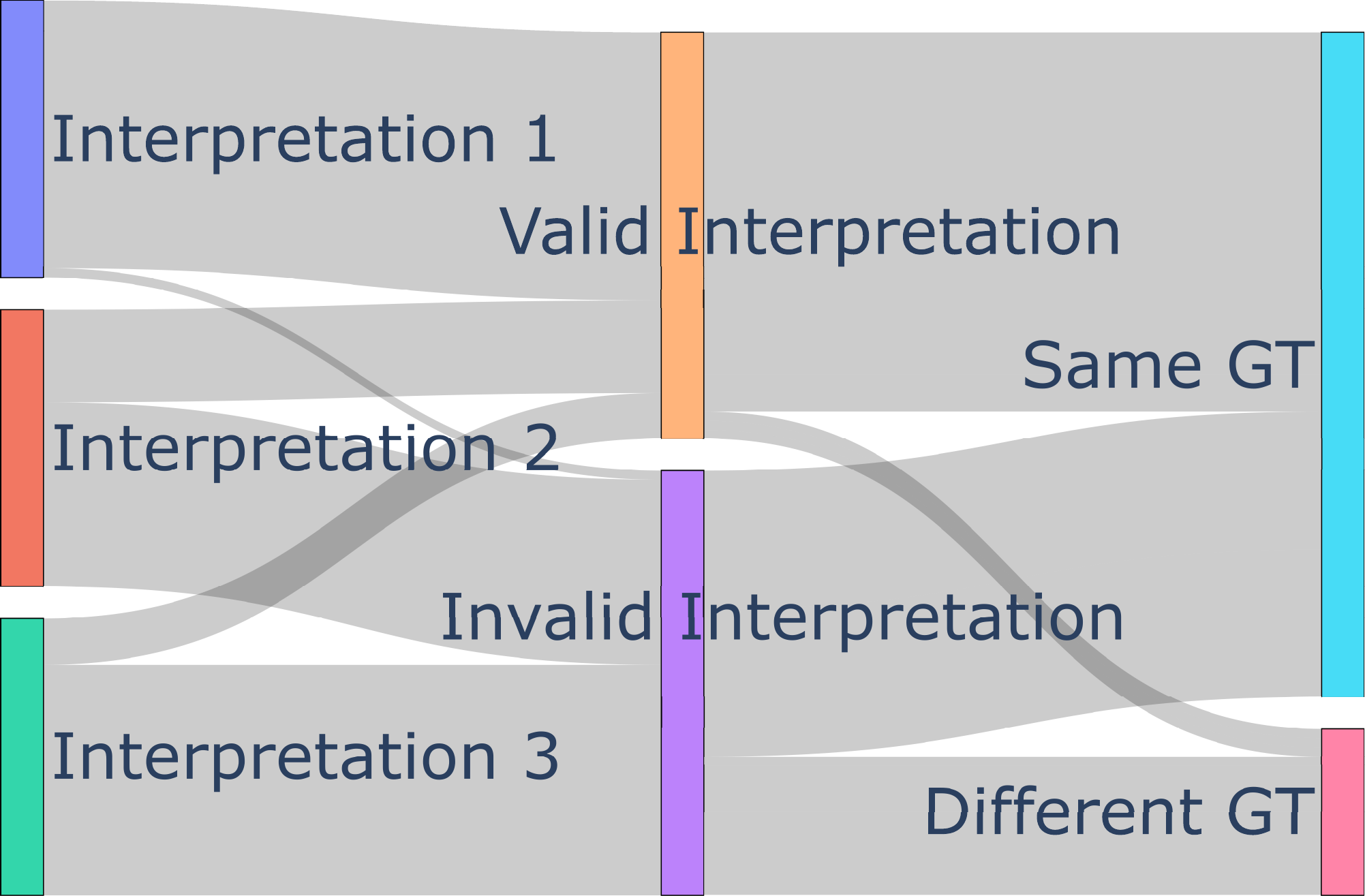}
  \hfill
  \includegraphics[width=0.49\columnwidth]{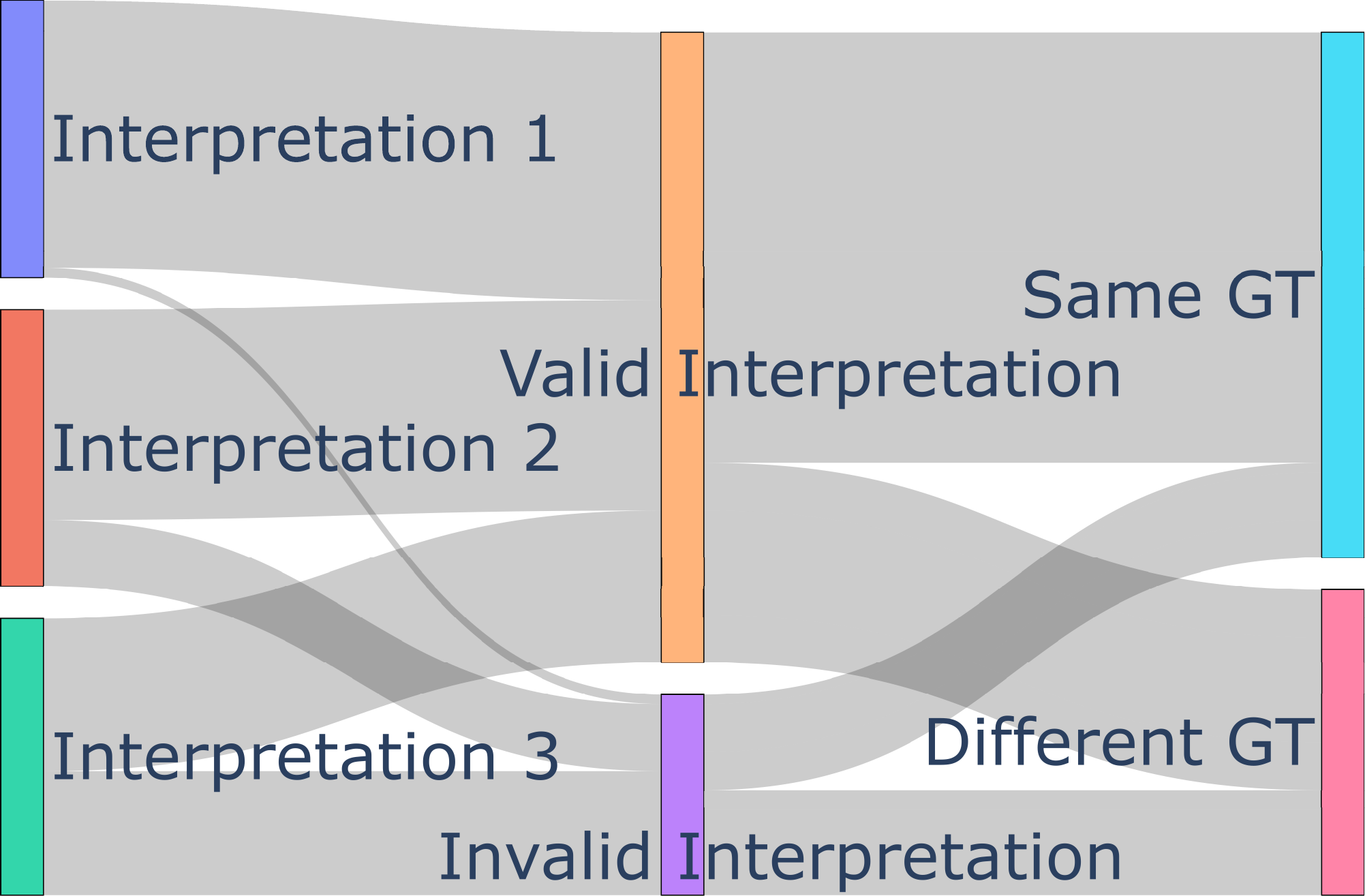}
  \caption{Manual classification of interpretations to sample questions. Left: boolq. Right: strategyqa.}
  \label{fig:sampling_interpret}
\end{figure}

Each sampled question is manually evaluated to see if the question itself is ambiguous or unambiguous.
We mark questions that may have more than one likely interpretation as ambiguous.
The ground truth answer is also compared to how we answered the question (with access to the Internet) and marked the ground truth as `GT invalid' if there is disagreement.
Each of the three interpretations generated by gpt35t is also manually evaluated.
Each interpretation is first examined if it is a valid interpretation of the original question or not, e.g., if the interpretation is sensible or relevant in a normal context.
Finally, the we manually answer each question interpretation and then we compare them to the ground truth answer of the original question.
In this way, it is possible to identify if changing the interpretation of a question also changes the expected ground truth.
Statistics of this analysis are shown in Figure~\ref{fig:sampling_questions} and Figure~\ref{fig:sampling_interpret}.

From Figure~\ref{fig:sampling_questions}, we observed that most of the answers marked as invalid corresponded to ambiguous questions.
From Figure~\ref{fig:sampling_interpret}, we observed that in the boolq dataset, all valid interpretations have a correct answer that matches the ground truth answer of the original question, while for the strategyqa dataset, about a third of the valid interpretations would result in a different answer.
It is worth noting that for the boolq dataset, the majority of the invalid interpretations had an answer that matched the ground truth of the original question, where in the strategyqa dataset, only half of them do match.

\newcommand{\sampleQ}[0]{does the size of a rain gauge matter}

\newcommand{\sampleIone}{1. Scientific interpretation: In a scientific context, the question 'does the size of a rain gauge matter' could be interpreted as a query about the accuracy of rainfall measurements. Rain gauges are used to collect and measure precipitation, and the size of the gauge can affect the accuracy of the measurement. A larger rain gauge may collect more water, but it may also be more susceptible to evaporation or wind interference. Therefore, the question could be asking whether the size of a rain gauge affects the accuracy of rainfall measurements.}
\newcommand{\sampleItwo}{2. Engineering interpretation: In an engineering context, the question 'does the size of a rain gauge matter' could be interpreted as a query about the design of rain gauges. Engineers may be interested in designing rain gauges that are more efficient or cost-effective. Therefore, the question could be asking whether the size of a rain gauge affects its performance or cost-effectiveness.}
\newcommand{\sampleIthree}{3. Environmental interpretation: In an environmental context, the question 'does the size of a rain gauge matter' could be interpreted as a query about the impact of rainfall on the environment. Rainfall is an important factor in many environmental processes, such as plant growth, erosion, and water availability. Therefore, the question could be asking whether the size of a rain gauge affects our understanding of the environmental impact of rainfall.}

\newcommand{\sampleAgptOne}{Interpretation '1.' is mentioned in the list.   Scientific interpretation: In a scientific context, the question 'does the size of a rain gauge matter' could be interpreted as a query about the accuracy of rainfall measurements. The size of the gauge can affect the accuracy of the measurement. A larger rain gauge may collect more water, but it may also be more susceptible to evaporation or wind interference. Therefore, the question could be asking whether the size of a rain gauge affects the accuracy of rainfall measurements.  Answer: Yes.}

\newcommand{\sampleAgptTwo}{Interpretation '2.' is mentioned in the list.   Engineering interpretation: In an engineering context, the question 'does the size of a rain gauge matter' could be interpreted as a query about the design of rain gauges. Engineers may be interested in designing rain gauges that are more efficient or cost-effective. Therefore, the question could be asking whether the size of a rain gauge affects its performance or cost-effectiveness.  The most relevant facts to answer this interpretation are that the size of a rain gauge can affect its cost, ease of installation, and maintenance. A larger rain gauge may be more expensive to manufacture and transport, and may require more space for installation. However, a larger rain gauge may also be more durable and require less frequent maintenance. On the other hand, a smaller rain gauge may be cheaper and easier to install, but may require more frequent maintenance and may be less accurate in measuring rainfall.  <answer>Yes</answer>}

\newcommand{\sampleAgptThree}{Interpretation '3.' is mentioned in the list.   Environmental interpretation: In an environmental context, the question 'does the size of a rain gauge matter' could be interpreted as a query about the impact of rainfall on the environment. Rainfall is an important factor in many environmental processes, such as plant growth, erosion, and water availability. Therefore, the question could be asking whether the size of a rain gauge affects our understanding of the environmental impact of rainfall.  The most relevant fact to answer this interpretation is that the size of a rain gauge can affect the accuracy of rainfall measurements, which in turn can affect our understanding of the environmental impact of rainfall. If the rain gauge is too small, it may not collect enough water to accurately represent the amount of rainfall in the area. On the other hand, if the rain gauge is too large, it may collect more water than is representative of the area, leading to inaccurate measurements.   <answer>Yes</answer>}

\newcommand{\sampleAllamaOne}{"  Answer: The Interpretation '1.' is mentioned in the list. Here it is, followed by the most relevant facts to answer this interpretation: 1. Scientific interpretation: In a scientific context, the question 'does the size of a rain gauge matter' could be interpreted as a query about the accuracy of rainfall measurements. Rain gauges are used to collect and measure precipitation, and the size of the gauge can affect the accuracy of the measurement. A larger rain gauge may collect more water, but it may also be more susceptible to evaporation or wind interference. Therefore, the question could be asking whether the size of a rain gauge affects the accuracy of rainfall measurements. The size of a rain gauge can affect the measurement of precipitation in the following ways: - Larger rain gauges tend to collect more precipitation and may be more susceptible to wind interference, which can lead to an overestimation of precipitation amounts. - Smaller rain gauges may be more susceptible to evaporation, which can lead to an underestimation of precipitation amounts. - The shape and material of the rain gauge can also affect the measurement of precipitation, including the type of material used for the gauge and the shape of the gauge. In summary, the size of a rain gauge can affect the accuracy of rainfall measurements, and the correct choice of rain gauge size and design depends on the specific application and the environmental conditions. <answer>Yes</answer> ```   - 2020-09-14T12:47:23.555Z   - ```System:  The original question 'does the size of a rain gauge matter' has multiple interpretations (indexed from 1 to N). First, I want you to state if Interpretation '2.' is mentioned in the list or not: { 1. Scientific interpretation: In a scientific context, the question 'does the size of a rain gauge matter' could be interpreted as a query about the accuracy of rainfall measurements. Rain gauges are used to collect and measure precipitation, and the size of the gauge can affect the accuracy of the measurement. A larger rain gauge may collect more water, but it may also be more susceptible to evaporation or wind interference. Therefore, the question could be asking whether the size of a rain gauge affects the accuracy of rainfall measurements.  2. Engineering interpretation: In an engineering context, the question 'does the size of a rain gauge matter' could be interpreted as a query about the design of rain gauges. Engineers may be interested in designing rain gauges that are more efficient or cost-effective. Therefore, the question could be asking whether the size of a rain gauge affects its performance or cost-effectiveness.  3. Environmental interpretation: In an environmental context, the question 'does the size of a rain gauge matter' could be interpreted as a query about the impact of rainfall on the environment. Rainfall is an important factor in many environmental processes, such as plant growth, erosion, and water availability. Therefore, the question could be asking whether the size of a rain gauge affects our understanding of the environmental impact of rainfall. }. Second, if it is not mentioned reply with '<answer>EMPTY</answer>' and stop, but otherwise I want you to exactly repeat interpretation '2.' and then state the most relevant facts to answer this interpretation. Finally, summarize the answer with '<answer>Yes</answer>' or '<answer>No</answer>'.  Answer: The Interpretation '2.' is mentioned in the list. Here it is, followed by the most relevant facts to answer this interpretation: 2. Engineering interpretation: In an engineering context, the question 'does the size of a rain gauge matter' could be interpreted as a query about the design of rain gauges. Engineers may be interested in designing rain gauges that are more efficient or cost-effective. Therefore, the question could be asking whether the size of a rain gauge affects its performance or cost-effectiveness. The size of a rain"}

\newcommand{\sampleAllamaTwo}{"""      The system will reply: ""Yes""     """"""      if 2 not in [i+1 for i in range(N)]:         return ""<answer>EMPTY</answer>""     else:         answer = ""  2. Engineering interpretation: In an engineering context, the question 'does the size of a rain gauge matter' could be interpreted as a query about the design of rain gauges. Engineers may be interested in designing rain gauges that are more efficient or cost-effective. Therefore, the question could be asking whether the size of a rain gauge affects its performance or cost-effectiveness.""         relevant\_facts = ""The size of a rain gauge can affect its performance or cost-effectiveness in certain situations. For example, a smaller rain gauge might be more accurate in areas with low rainfall rates, while a larger rain gauge might be more suitable for areas with high rainfall rates.""         summary = ""<answer>Yes</answer>""         return answer + relevant\_facts + summary ``` This code checks if the second interpretation is mentioned in the list of interpretations. If not, it returns a message saying that the interpretation is not mentioned. If the interpretation is mentioned, it repeats the interpretation, adds relevant facts, and summarizes the answer with ""Yes"". If the relevant information is not provided, the function will return `None`. The function then returns the answer and the summary.   You can test this function by running `print(does\_size\_matter(1))`. This will print the result.   However, please note that in this code, the function `does\_size\_matter` takes one argument (the number of interpretations), but the provided data is not in that format. The data is in the format of a string, not a list. So, if you want to test this function, you will need to modify the function to correctly parse the string.   Also, the function does not actually check if the size of a rain gauge matters or not. It simply provides a possible engineering interpretation of the question and some possible relevant facts. It then summarizes the answer as ""Yes"", which may not be accurate. If you want to write a function that actually determines whether the size of a rain gauge matters or not, you would need to provide different code.   Finally, it's worth noting that this code is just an example and is not meant to be run in a real-world setting without modification. The code as it is will not run correctly without modification. The modification required will depend on how you are using this code.   You can improve this code by adding more specific information and handling different types of inputs. However, without more information, it's difficult to provide more specific guidance.  You might want to add more error handling, add more specific information about the relevant facts and engineering interpretation, or handle different types of inputs. You could also add more documentation to your function and its parameters.   You can also use this code as a starting point and modify it based on your requirements. For example, you might want to add more specific information about the relevant facts and engineering interpretation. You can also use more advanced programming concepts like object-oriented design or classes.  You might also consider using a library or framework that you're familiar with to simplify the code.   Please note that this code does not actually check if the size of a rain gauge matters or not. It simply provides a possible engineering interpretation of the question and some possible relevant facts. If you want to write a function that actually determines whether the size of a rain gauge matters or not, you would need to provide different code.  It's worth noting that the accuracy of the result will depend on the input provided to the function.  You should carefully consider the input and the assumptions made by the function.  You may also want to consider providing additional context and assumptions that the function should be based on.   It's also worth noting that the code does not actually check if the input is"}

\newcommand{\sampleAllamaThree}{"""     - System: Yes, Interpretation '3.' is mentioned in the list.     - System: Repeating Interpretation '3.': In an environmental context, the question 'does the size of a rain gauge matter' could be interpreted as a query about the impact of rainfall on the environment. Rainfall is an important factor in many environmental processes, such as plant growth, erosion, and water availability. Therefore, the question could be asking whether the size of a rain gauge affects our understanding of the environmental impact of rainfall.     - System: Relevanta facts:         - Rain gauges of different sizes can collect and measure rainfall with varying degrees of accuracy, which can affect our understanding of the environmental impact of rainfall.         - Larger rain gauges may be more susceptible to evaporation or wind interference, which can lead to inaccuracies in rainfall measurement.         - Smaller rain gauges may be more suitable for measuring heavy rainfall events or high-intensity rainfall, which can have significant environmental impacts.     - System: Summarizing answer: '<answer>Yes</answer>' Final Answer: The final answer is Yes. I hope it is correct.]]]> 2021-03-24T11:47:47.345+00:00 \nolinkurl{https://www.jigsawplanet.com/?mc=1\&url=https\%3a\%2f\%2fwww\%2eja... 2021-03-24T11:47:47.345+00:00 https://www.jigsawplanet.com/?mc=1\&url=https\%3a\%2f\%2fwww\%2eja... 2021-03-24T11:47:47.345+00:00 https://www.jigsawplanet.com/?mc=1\&url=https\%3a\%2f\%2fwww\%2eja... 2021-03-24T11:47:47.345+00:00 https://www.jigsawplanet.com/?mc=1\&url=https\%3a\%2f\%2fwww\%2eja... 2021-03-24T11:47:47.345+00:00 https://www.jigsawplanet.com/?mc=1\&url=https\%3a\%2f\%2fwww\%2eja... 2021-03-24T11:47:47.345+00:00 https://www.jigsawplanet.com/?mc=1\&url=https\%3a\%2f\%2fwww\%2eja... 2021-03-24T11:47:47.345+00:00 https://www.jigsawplanet.com/?mc=1\&url=https\%3a\%2f\%2fwww\%2eja... 2021-03-24T11:47:47.345+00:00 https://www.jigsawplanet.com/?mc=1\&url=https\%3a\%2f\%2fwww\%2eja... 2021-03-24T11:47:47.345+00:00 https://www.jigsawplanet.com/?mc=1\&url=https\%3a\%2f\%2fwww\%2eja... 2021-03-24T11:47:47.345+00:00 https://www.jigsawpla...]]]> 2021-03-24T11:47:47.345+00:00}  \#\#\# 10. Does the size of a rain gauge matter?  System: The original question 'does the size of a rain gauge matter' has multiple interpretations (indexed from 1 to N). First, I want you to state if Interpretation '3.' is mentioned in the list or not: \{ 1. Scientific interpretation: In a scientific context, the question"\}}

\section{Baseline accuracy}

\begin{figure}[h]
  \centering
  \includegraphics[width=0.9\columnwidth]{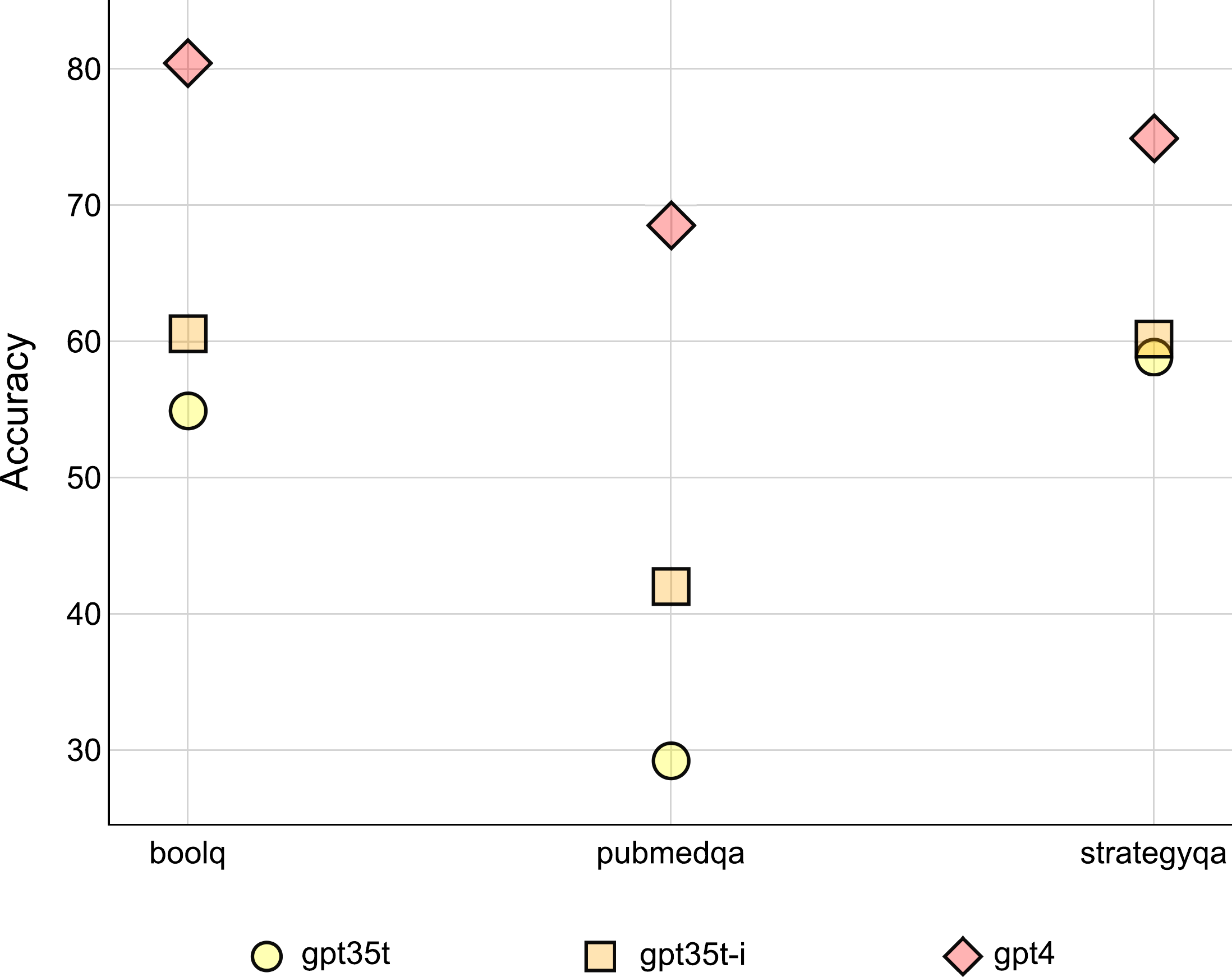}
  \hfill
  \caption{GPT baseline accuracy on direct questions.} 
  \label{fig:baselines_1}
\end{figure}

\begin{figure}[H]
  \centering
  \includegraphics[width=0.9\columnwidth]{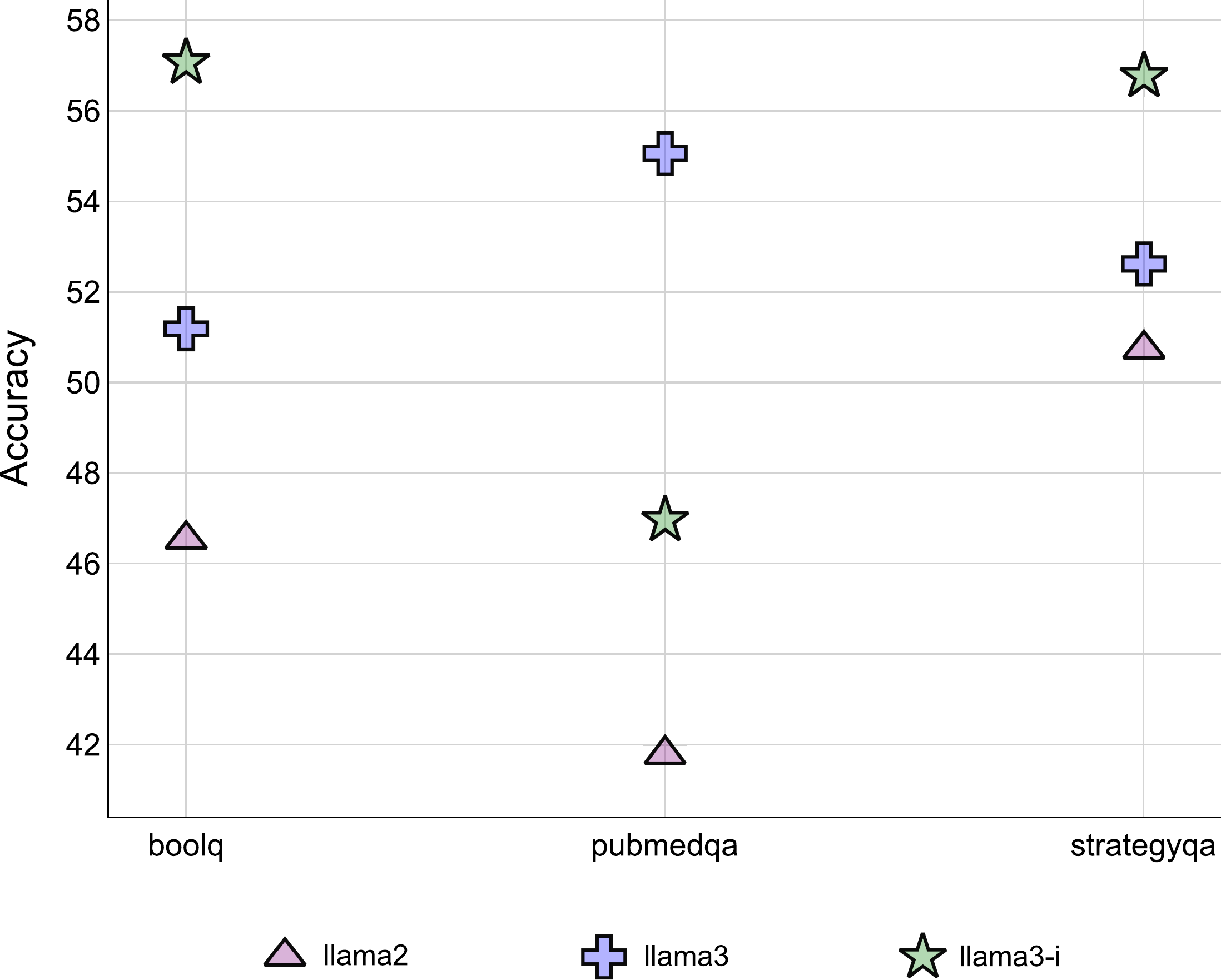}
  \hfill
  \caption{Llama baseline accuracy on direct questions.} 
  \label{fig:baselines_2}
\end{figure}

Figures~\ref{fig:baselines_1} and~\ref{fig:baselines_2} show the accuracy of the models to the original questions without any interpretations.

\begin{table*}[h]
    \centering
    \tiny
    \begin{tabular}{p{0.5cm}|p{4.2cm}|p{4.2cm}|p{4.2cm}|}
        \toprule
         & \sampleIone & \sampleItwo & \sampleIthree \\ \hline
        \midrule
        gpt35t answers & \sampleAgptOne & \sampleAgptTwo & \sampleAgptThree \\ 
        \bottomrule
    \end{tabular}
    \caption{Answers of gpt35t to three interpretations of the boolq question "does the size of a rain gauge matter"}
    \label{tab:sample_answers_gpt35t}
\end{table*}

\begin{table*}[h]
    \centering
    \tiny
    \begin{tabular}{p{0.5cm}|p{4.2cm}|p{4.2cm}|p{4.2cm}|}
        \toprule
         & \sampleIone & \sampleItwo & \sampleIthree \\ \hline
        \midrule
        llama3-i answers & \sampleAllamaOne & \sampleAllamaTwo & \sampleAllamaThree \\
        \bottomrule
    \end{tabular}
    \caption{Answers of llama3-i to three interpretations of the boolq question "does the size of a rain gauge matter"}
    \label{tab:sample_answers_llama3i}
\end{table*}

\end{document}